\documentclass{article}

\usepackage{arxiv}

\usepackage[utf8]{inputenc} 
\usepackage[T1]{fontenc}    
\usepackage{hyperref}       
\usepackage{url}            
\usepackage{booktabs}       
\usepackage{amsfonts}       
\usepackage{nicefrac}       
\usepackage{microtype}      
\usepackage{graphicx}
\usepackage{tgtermes}
\usepackage{mathptmx}  
\usepackage[scaled=.92]{helvet}
\usepackage{amsthm,amsmath,amssymb}
\usepackage{mathrsfs}
\usepackage{todonotes}
\usepackage[numbers]{natbib}  

\newtheorem{theorem}{Theorem}[section]

\newtheorem{claim}[theorem]{Claim}

\theoremstyle{definition}
\newtheorem{definition}[theorem]{Definition}

\theoremstyle{remark}

\graphicspath{{media/}}     

\title{The Variability of Model Specification
}

\author{
    Joseph R. Barr \\
    AI Group\\
    Acronis SCS\\
    Scottsdale, Arizona,
    United States of America\\
  \texttt{joe.barr@acronisscs.com} \\
   \And
    Peter Shaw\\
    Department of AI\\
    Nanjing Uni. of Information Science \& Technology\\
    JiangSu, China\\
   \texttt{peter.shaw.cs@gmail.com} \\
  \And 
    Marcus Sobel\\
    Department of Statistics\\
    Temple University \\
    Philadelphia, United States of America\\
    \texttt{marc.sobel@temple.edu}
}

\begin{document}
\maketitle

\begin{abstract}
It's regarded as an axiom that a good model is one that compromises between bias and variance. The bias is measured in training cost, while the variance of a (say, regression) model is measure by the cost associated with a validation set. If reducing bias is the goal, one will strive to fetch as complex a model as necessary, but complexity is invariably coupled with variance: greater complexity implies greater variance.  In practice, driving training cost to near zero does not pose a fundamental problem; in fact, a sufficiently complex decision tree is perfectly capable of driving training cost to zero; however, the problem is often with controlling the model's variance. 
We investigate various regression model frameworks, including generalized linear models, Cox proportional hazard models, ARMA, and illustrate how misspecifying a model affects the variance. 
\end{abstract}

\keywords{Statistical Models \and 
{Regression Models}
\and {Variance-Bias Tradeoff} \and {Model Misspecification} \and{Cauchy Eigenvalues Interlacing Theorem} \and {Spectral Radius of a Matrix} \and {Principle of parsimony} \and {Cox PH Regression} \and {ARMA}
}

\section{Introduction}
In recent years the machine learning approach to data analysis has supplanted the classical statistical dogma. Testing model's performance against a `hold-out' dataset has greatly displaced erstwhile common practice of testing hypotheses about model's parameters. Arguably, both approaches are valid, and both represent different facets of validating the aptitude of a statistical model \cite{Barr}. Methodologies also have evolved, which slant the approach to modeling in the direction of regularization. The \textsc{lasso} method has become a  de-facto replacement for testing $H_0: \beta = 0$ against $H_1: \beta \neq 0$; however, lasso goes much farther as it's commonly used to control the complexity of neural networks, etc. The tools which depict the \textsc{bias-variance tradeoff} have cast the classic \textsc{Ockham's Razor} into quantifiable statistical terms. Moreover, the authors feel that something was missing in the translation from the classical to the modern. By focusing on classical regression methodology, we strive to fill a few of the holes and put the principle of bias-variance tradeoff on solid mathematical foundations.  

\subsection{The Bias-Variance Tradeoff of a Regression Model}
\label{intro}
\begin{figure}[hb!]
    \centering
    \includegraphics[height=8cm]{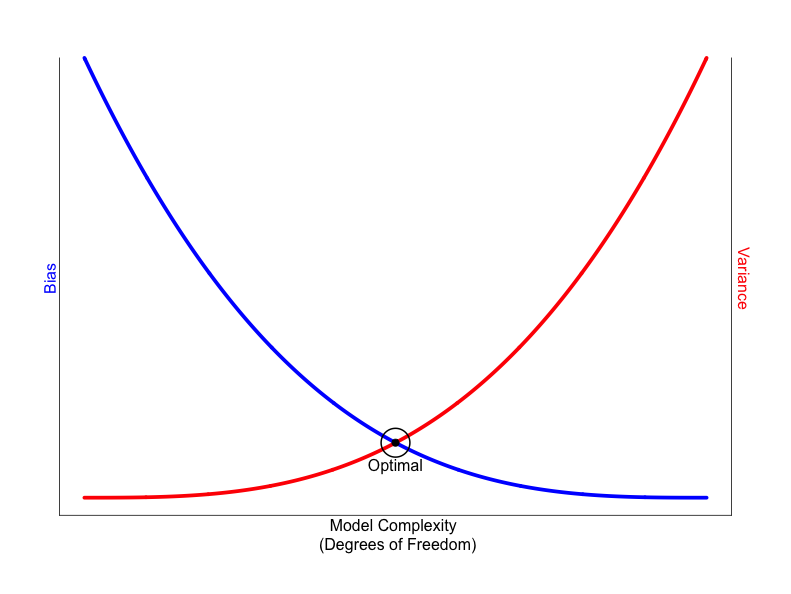}
    \caption{Bias-Variance Tradeoff
    \label{fig:my_label}}
\end{figure}

With a small sacrifice in generality, we consider the simplest generic regression models and two-part \textsc{train} \& \textsc{test} partitioning of the data.
\textsc{Training pairs} $\mathcal{D} =\{ (x^1, y^1), (x^2,y^2),...,(x^N,y^N)\}$ with $x^j \in \mathcal{X} \subset \mathbb{R}^d$, $y\in \mathcal{Y} \subset \mathbb{R}$ are split in two disjoint subsets $\mathcal{T}$ and $\mathcal{V}$, with $\mathcal{T}$ with $\mathcal{T}\cup \mathcal{V}=\mathcal{D}$. We take $|\mathcal{T}|=pN$ with say, $0.2 \leq p \leq 0.8$ and, of course, $|\mathcal{V} |=(1-p)N$.  We train a model $\phi:\mathcal{X} \rightarrow \mathcal{Y}$ on a subset $\mathcal{T}$ by driving the \textsc{cost} as low as possible.  The training cost $\mathcal{C}$ is a function of the \textsc{complexity} of the model; an increase in complexity generally decreases training cost. This may be parameterized  by a subscript, i.e., by writing $\mathcal{C}_n$ for the cost  of the model's \textsc{degrees of freedom} $n$, where the degrees of freedom is synonymous with complexity. To illustrate how complexity is related to cost, given $n$ points on the plane, there is a polynomial of degree $n-1$ that fits perfectly, i.e., the cost of training $\mathcal{C}=0$. This can be done using the Lagrange interpolation polynomial. Still, a lower degree polynomial sacrifices fitting any particular point in order to improve the overall fit. A happy consequence of fitting $n$ points with a polynomial of degree lower than $n-1$ is the expectation that test (or new) data will deviate less from model values. The general \textsc{bias-variance} phenomenon is captured by a rather idealistic bias-variance tradeoff depicted in Fig. \ref{fig:my_label}.


Although it is strongly believed that for the two graphs, the bias and variance meet at an optimal point where model complexity is `just right', that optimum is often difficult to calculate; it's typically estimated empirically.

\section{The Specification of a Linear Model}\label{LM}
We'll use standard notation found in, e.g. Christensen `Plane Answers to Complex Problems'  \cite{Christensen2011}, and without loss of generality, consider a zero-intercept linear model $$Y= X\beta+\epsilon $$ where $Y\in \mathbb{R}^N$ is an observable vector, $X\in \mathbb{R}^{N\times p}$ is a matrix of known quantities, $\beta \in \mathbb{R}^p$ an unobservable vector, and $\epsilon \in \mathbb{R}^N$ is an unobservable vector. We also assume that $p< N$, that $X$ has more rows than columns, and that rank$(X)=p$. It's common to summarize a linear model and write $$Y \sim \mathcal{N}_p(X\beta, \sigma^2 I_N) $$ with $\beta \in \mathbb{R}^p$, $\sigma^2>0$ unknown parameters, $I_N$ is the $N\times N$ identity matrix, and $\mathcal{N}_p$ is a $p$-dimensional normal (Gaussian) random variable.

A  modeler's task is to select an optimal subset of the columns of $X$. Much has been written on ``subset selection''. Since Fisher's F-test, \cite{Christensen2011}
many algorithms were developed to select an optimal subset of features. Those include eliminating insignificant coefficients ($\beta=0$) using Fisher $F$-test, feature subset selection via minimization of Akaike Information Criterion (AIC) \& Bayes Information criterion (BIC) [Cavanaugh, 2019] \cite{cavanaugh2019akaike}. Feature elimination to remove collinearity with the variance inflation factor (VIF) [Lin, et al., 2011] has also been used to select optimal subsets of features \cite{lin2011vif}.

Notwithstanding all those techniques, the problem of variable selection is currently addressed with regularization, primarily a mix of lasso $(\mathcal{L}^1)$ and ridge $(\mathcal{L}^2)$. For variable selection, lasso shrinkage is particularly effective [Tibishani, 1997] \cite{tibshirani1996regression} and is used in various modeling techniques, from generalized linear models to neural networks; regularization is regarded as an all-purpose technique to control model complexity. Grid search is a reliable performer to balancing bias and variance. Needless to say, to produce an adequate linear model, appropriate steps to ensure data is apt for a linear model must be taken. These include  outlier removal and  feature transformation, etc.., but for the purpose of this discussion, data treatment won't be included.

\textsc{Estimation}. Under the assumption that rank$(X)=p$, the symmetric $p\times p $ matrix $X^TX$ is non-singular; therefore, the inverse $(X^TX)^{-1}$ exists. The \textsc{ordinary least squares (OLS)} estimate is $\widehat{\beta } = (X^TX)^{-1} X^TY$. By the Gauss-Markov Theorem, restricted to linear models, the unbiased estimate having the lowest variance and optimal OLS estimates are equivalent. The estimate $\widehat{Y} = X\widehat{\beta }$ is an orthogonal projection of $Y$ onto the column space of $X$. The variance of $\widehat{\beta}$ is $$
var(\widehat{\beta}) = \sigma^2 (X^TX)^{-1}
$$
and the estimate of $\sigma^2 $ is 
$$
\widehat{\sigma^2} = \dfrac{Y^T(I-(X^TX)^{-1}X^T)Y}{N-p}.
$$

\subsection{Model Nesting}
The model $Y=X\beta + \epsilon$ is, by definition, the \textit{full or saturated model}. For a matrix $X^*$ consisting of a subset of the columns of $X$, say $q$ columns, $q<p$, the model $Y=X^*\beta^* + \epsilon$ is called a \textit{nested model}. The unobserved vector $\beta^* \in \mathbb{R}^q$, (corresponding to the nested model) has fewer parameters than the saturated model. So, the linear model $Y=X^*\beta^* + \epsilon$ has lower complexity than the saturated linear model  $Y=X\beta + \epsilon$. 

\section{Bias-Variance in Linear Models} 
Consider the saturated and nested linear models $Y=X\beta+\epsilon$, $Y=X^*\beta^* + \epsilon$, respectively.  As always, we assume that the matrix $X$ is full rank, and a fortiori, so is $X^*$. The \textsc{bias} of the model $Y=X\beta+\epsilon$ is measured against MSE $$
bias(Saturated) = \min_{\beta \in \mathbb{R}^p} \|Y-X\beta\|^2
$$
and the bias of the nested model $Y = X^* \beta^* + \epsilon$ is 
$$
bias(Nested) = \min_{\beta^*\in \mathbb{R}^q} \| Y-X^* \beta^*\|^2.
$$
Since the saturated model is minimized over a larger set of parameters, $$
bias(Saturated) \leq bias(Nested).
$$
The claim we wish to establish is that the variance of the saturated model is always at least as large as the variance of the nested model.

\subsection{The variance of the Linear Model} 
The variance of a linear model $Y=X\beta+\epsilon$ is a matrix $var(\widehat{\beta})=\sigma^2 (X^TX)^{-1}$. Recall that an unbiased estimate of $\sigma^2$ is 
$$\widehat{\sigma^2} = \dfrac{Y^T(I-X(X^TX)^{-1}X^T)Y}{N-p}.$$
Given a saturated linear models $Y=X\beta+\epsilon$ and a nested linear model $Y=X^*\beta^* + \epsilon$, we wish to establish which of the two has the lower variance.

The bias-variance tradeoff stipulates that the less complex model has lower variance, and this is precisely what we will prove. We demonstrate how it follows from the celebrated \textsc{Cauchy’s Eigenvalues Interlacing Theorem},  which we state without a proof. An accessible proof is found in [Huang, 2004] \cite{hwang2004cauchy}.
\vspace{5pt}
\begin{definition}
Let $A$ be a square $n\times n$ matrix. The $(n-1)\times (n-1)$ matrix $A^*$ is a \textsc{principal} submatrix of $A$ if $A^*$ is obtained from $A$ by removing row $i_k$ and column $i_k$. Similarly, $A^*$ is a $q\times q$ principal submatrix of $A$ is $A^*$ is obtained by removing $q$ rows and columns $i_1<i_1 < \dots i_q$.
\end{definition}
\vspace{5pt}
\begin{theorem} (Cauchy's Eigenvalue Interlacing Theorem) Let $A$ be a non-singular real symmetric $n\times n$ matrix and let $A^*$ be a principal $(n-1)\times(n-1) $ submatrix consisting of the first $n-1$ rows and $n-1$ columns of $A.$ Let $\lambda_1 \geq \lambda_2 \geq \dots \geq \lambda_n>0$ be the eigenvalues of $A$ and $\mu_1 \geq \mu_2 \geq \dots \geq \mu_{n-1}$ the eigenvalues of $A^*$. Then     $\lambda_1 \geq \mu_1 \geq \lambda_2 \geq \mu_2 \geq \dots \geq \mu_{n-1} \geq \lambda_n$. 
$\square$ 
\end{theorem}

\noindent
Note that there's nothing special about which row and column of $A$ are omitted because two matrices which differ by a simultaneous row and column permutation are similar, hence have equal spectra.

A consequence of the theorem is that the eigenvalues of the inverses $A^{-1}$ and $(A^*)^{-1}$ are interlaced 
$$
\lambda_n^{-1} \geq \mu_{n-1}^{-1} \geq \lambda_{n-1}^{-1} \geq \mu_{n-2}^{n-1} \geq \dots \geq \mu_1^{-1} \geq \lambda_1^{-1}.
$$
\vspace{5pt}
Another immediate consequence which we plan to use is:
\begin{theorem}
Let $A$ be $n\times n$ real symmetric matrix with spectral radius $\lambda_1$ and $A^*$ a principal $q\times q$ symmetric principal submatrix of $A$ ($q<n$) with spectral radius $\lambda^*_1$. Then $\lambda_1^* \leq \lambda_1$.  $\square$
\end{theorem}
 
\textbf{Notation for largest eigenvalue, spectral radius:} As a matter of expediency, we'll write $\lambda_1(A)$ or just $\lambda_1$, if the matrix is understood from the context, to denote the largest eigenvalue of $A$, i.e., the \textsc{spectral radius} of $A$.

As we saw earlier, if $\lambda_n$ is the smallest eigenvalue of a  non-singular real symmetric $A$, then $\lambda_1(A^{-1}) = \dfrac{1}{\lambda_n}$. We trust that this rotational standard won't cause any confusion.

Since we'll be dealing exclusively with non-singular real symmetric matrices, we don't need to worry about absolute values.

\section{Comparing variances of nested linear models}
Given a saturated linear models  $Y=X\beta+\epsilon$, where $X \in \mathbb{R}^{N\times p}$, and $\epsilon \sim N(0, \sigma^2 I)$. And a nested linear model 
$Y=X^*\beta^* + \epsilon^{*}$, where $X^{*} \in \mathbb{R}^{N\times q}$, and $q<p$, and  $\epsilon^{*} \sim N(0, {\sigma^{*}}^{2} I) $.  
We solve the problem in two steps. We first analyze the magnitudes of the estimate of the variance of the residual for saturated and nested models. $(N-p)\widehat{\sigma^2}$ of the saturated model is the magnitude of the length of the residual of the saturated model $(I-X(X^TX)^{-1}X^T)Y$, while $(N-q)\widehat{{\sigma^{*}}}^{2}$  is the length of the residual vector $(I-{X^*}{({X^*}^TX^*)}^{-1}{X^*}^T)Y$. To simplify notation, put $M=X(X^TX)^{-1}X^T$ and $M^*=X^{*}({X^{*}}^{T}X^{*})X^{*}$ where $M$ is a projection on the column space of $X$ and $M^{*}$ is the projection on the column space of $X^{*}$.

\vspace{5pt}
\begin{claim}
$$
\|(I-M)Y \| \leq \|(I-M^*)Y\|
$$
\end{claim}

To convince oneself of the veracity of the claim, one notices that a projection of a point $Y$ onto the column space $C(X)$ of $X$ is the closest point on the convex set $C(X)$ and is closer than the point closest to a convex subset $C(X^*)$ of $C(X)$.  Since $q<p$, $\dfrac{1}{N-p} > \dfrac{1}{N-q}$. 
Consequently, the two estimates $\widehat{{\sigma}^2} $ and $\widehat{{\sigma^*}^2} $ are 

$$
\widehat{\sigma^2} = \dfrac{Y^T(1-M)Y}{N-p}, 
$$
and 
$$
\widehat{{\sigma^*}^2} = \dfrac{Y^T(1-M^*)Y}{N-q} 
$$
We proceed with heuristics. In most cases, $p$ is much smaller than $N$, where $N$ is more than ten times bigger than $p$, often a hundred times bigger. So for all intent and purpose $\dfrac{1}{N-p} \approx \dfrac{1}{N-q} \approx \dfrac{1}{N}$. Therefore, $\widehat{\sigma^2} \approx \dfrac{1}{N} Y^T (I-M)Y$ and $\widehat{{\sigma^*}^2} \approx \dfrac{1}{N} Y^T(I-M^*)Y$ and based on this heuristics,  $\widehat{\sigma^2} \leq \widehat{{\sigma^*}^2}$.  

The second step is estimating the magnitudes of the matrices $(X^TX)^{-1}$ and $({X^*}^TX^*)^{-1}$. 

Recall that for a symmetric matrix $A$, the magnitude of $A$ is commonly measured by its spectral radius, i.e., its largest eigenvalue. To compare the spectral radii of ${(X^TX)}^{-1}$ and ${({X^*}^TX^*)}^{-1}$, we use Cauchy's eigenvalues interlacing theorem.
Because the symmetric matrix ${X^*}^TX^*$ is a principal submatrix of the symmetric non-singular matrix $X^TX$ it follows that their eigenvalues interlace. In particular, the smallest eigenvalue $\lambda_p$  of ${(X^TX)}^{-1}$ is smaller than the smallest eigenvalue $\mu_q$ of ${({X^*}^TX^*)}^{-1}$. Also,  non-singularity of $X^TX$ implies $\lambda_p>0$, and so $\mu_q > \lambda_p>0$. Therefore, $\dfrac{1}{\lambda_p} > \dfrac{1}{\mu_q}$. This means that the spectral radius of ${({X^*}^TX^*)}^{-1}$ is smaller than the spectral radius of ${(X^TX)}^{-1}$. Combining with what we saw earlier, $var(\widehat{Y})$ of the saturated model is larger than $var(\widehat{Y})$ of the nested model.

\noindent$\square$

To summarize, we have proved the theorem:

\begin{theorem}
Consider a saturated model $L$ and a nested model $L^*$. Then 
$$
bias(L) \leq bias(L^*)
$$
and $$
var(L) \geq var(L^*)
$$
$\square$
\end{theorem}

\section{Generalized Linear Model}
A \textsc{generalized linear model (GLM)} is based on observations $Y_j \sim \mathcal{P}(\eta_j \vert \beta_j)$ where $\mathcal{P}$ is of an exponential-type distribution, with weight vector $w_j$, and $\eta_j = \beta^T x_j$. 
The mean $\mu_j= E(Y_j \vert X_j)=h(X_j^T\beta)$ depends on $X_j$ via a smooth invertable link function $h$ with $\mu_j=h(\eta_j)$ or $g(\mu_j)=h^{-1}(\mu_j) = \eta_j$. The three most important generalized linear models are the:
\begin{enumerate}
    \item Gaussian: 
    \[ p(Y_j; \mu,\sigma^2) = \dfrac{1}{\sqrt{2\pi \sigma^2}}\exp\Big(\dfrac{-(Y_j-\mu)^2}{2\sigma^2} \Big) \]
    with link function $h(t)=t$ and $\mu=X_j'\beta$.
    \item Binomial: with $Y=Y_1+...+Y_n$ and 
    \[ p(Y_j \vert \pi_j)=  \pi_j^{Y_j} (1-\pi_j)^{1-Y_j}: j=1,...,n \]
    and link function:
    \[ \pi_j = {\rm logit}^{-1}(X_j'\beta) = h(X_j^T\beta) \]
    \item Poisson: 
    \[ f(Y_j; \mu_j)=  \dfrac{\mu_j^{y_j}}{y_j!}\exp(-\mu_j) \]
    and link function
    \[ \mu_j = \exp \left( X_j^T \beta \right) = h(X_j^T \beta) \]
\end{enumerate}
 The parameters $\beta=(\beta_1,...,\beta_p)$ are estimated by maximizing the likelihood of the observations as a function of $\beta$  (\textsc{MLE}).  Classic central limit theorems show that, for the saturated model, the Fisher matrix (whose inverse is given by the Fisher information) is:
 \begin{equation} F(\mathbf{X})= \mathbf{X}^T \mathbf{W} \mathbf{X} \label{satGLM} \end{equation}
 where $X$ is the matrix comprising of the $p$ columns $X_1,...,X_p$
 \[ 
 \mathbf{X}= {\rm 
 \begin{bmatrix}
  | & |  & . & . & | \\
 X_1 & X_2 & . & . & X_p \\
   | & |   & . & . & | 
 \end{bmatrix}
 } \]
 and $\mathbf{W}$ is the diagonal weight matrix. 
 \[ \mathbf{W} = {\rm diag}(W_{i,i}) = \frac{(h(\eta_i))^2}{\sigma_i^2} \]
 Recall that $\eta_i=g(\mu_i)$ $(i=1,...,p)$ and
 \[ \sigma_i^2 = Var(Y_i) \]
 
 For the nested model, we assume a subset $\mathbf{X}^*$ of the full set $\mathbf{X}$ of covariates. The resulting nested Fisher matrix is:
 \begin{equation} F(\mathbf{X}^*)= [\mathbf{X}^*]^T \mathbf{W^*} \mathbf{X}^* \label{nestGLM} \end{equation}
 where $X^*$ is composed of $X_1^*, ..., X_m^*$, a subset of  the columns of $X$:
 \[ 
 \mathbf{X^*}= {\rm 
 \begin{bmatrix}
  | & |  & . & . & | \\
 X_1^* & X_2^* & . & . & X_m^* \\
   | & |   & . & . & | 
 \end{bmatrix}
 } \]
 and 
 \[ \mathbf{W^*} = {\rm diag}(W_{i,i}^*), \]
${i=1,..,m}$,  is a submatrix of $\mathbf{W}$.   Under standard assumptions about the weight matrix, Cauchy's theorem shows that, using (\ref{satGLM},\ref{nestGLM}):
 \begin{equation}    
 \lambda_1\big( F(\mathbf{X}^*)\big) \leq \lambda_1\big(F(\mathbf{X})\big) \label{step} \end{equation}
 and we can infer from (\ref{step}) that
 \begin{equation} 
 \lambda_1\big( F^{-1}(\mathbf{X}^*) \big) \geq \lambda_1\big( F^{-1}(\mathbf{X})\big) \label{end} \end{equation}
 This demonstrates our result.

\subsection{Cox Proportional Hazards Model}
The Cox proportional hazards model \cite{Cox1984}, \cite{Cox} assumes covariate parameters: 
\[ \beta=(\beta_1,...,\beta_p) \],
realized values of the covariates for subject $i$;
\[ X_i= (X_{i1},...,X_{ip}) \]
($i=1,...,n$), and link functions $\theta_i=\exp(X_i^T \beta)$.
We assume independent censoring variables: 
\[ C_1,...,C_T \] 
(for times $t_1<...<t_T$) with $C=1$ denoting the lack of censoring).
It is assumed that the number of subjects $n$ is asymptotically large. 
  As shown by \cite{Vaart} and  \cite{Efron}, inference about $\beta$
  can be based entirely on the (saturated) log partial likelihood, which takes the form:
  \begin{equation} l(\beta)= \sum_{i: C_i=1} \left\{ X_i^T \beta - \log \left( \sum_{t_j \geq t_i} \exp(X_j^T\beta) \right) \right\} \label{sat} \end{equation}
 A nested model takes the form:
  \begin{equation} l^*(\beta)= \sum_{i: C_i=1} \left\{ [X_i^*]^T \beta^* - \log \left( \sum_{t_j \geq t_i} \exp([X_j^*]^T \beta^*) \right) \right\} \label{next} \end{equation}
  for column vectors $\{ X^* \} \subset \{X \}$ contained in the saturated model.  The variances of $\beta$ in equation (\ref{sat}) and $\beta^*$ in equation (\ref{next}) are based on the inverses of the partial likelihood second derivatives (i.e., fisher information):
  \begin{eqnarray} F(\beta) & = & \sum_{C_i=1} \left( \frac{\sum_{j:t_j \geq t_i} \exp(X_j^T \beta)X_j X_j^T}{
  \sum_{j: t_j \geq t_i} \exp(X_j^T\beta)} 
   - \frac{\sum_{t_j \geq t_i} \exp(X_j^T\beta) X_j \sum_{t_j \geq t_i} \exp(X_j^T \beta) X_j^T}{ \left(\sum_{t_j \geq t_i} \exp(X_j^T\beta) \right)^2} \right) \nonumber \end{eqnarray}
   \begin{eqnarray} F(\beta^*) & = & \sum_{C_i=1} \left( \frac{\sum_{j:t_j \geq t_i} \exp(X_j^T \beta^*)X_j X_j^T}{
  \sum_{j: t_j \geq t_i} \exp(X_j^T\beta^*)} 
   - \frac{\sum_{t_j \geq t_i} \exp(X_j^T\beta^*) X_j \sum_{t_j \geq t_i} \exp(X_j^T\beta^*) X_j^T}{ \left(\sum_{t_j \geq t_i} \exp(X_j^T\beta^*) \right)^2} \right) \nonumber \end{eqnarray}
   Thus, using the `hat' notation $ ` \widehat{} ' $ for the MLE, our main result suggests that 
   \[
    \lambda_1\big(F^{-1}(\widehat{\beta}^*)\big) \leq
    \lambda_1\big(F^{-1}(\widehat{\beta})\big),
   \]
   i.e., the largest eigenvalue of the nested variance is smaller than that of the saturated model.

\subsection{Time Series} 
\vspace{1pt}
Time series data consists of time-stamped observations $x_t$, with $t\in \mathbb{Z}$, i.e., the data is a `two-tailed' series $(\dots, x_{-3}, x_{-2}, x_{-1}, x_0, x_1, x_2, x_3, \dots )$.  We assume a large number $n$ of such time-stamped observations. This gives rise to accurate estimates $\widehat{\gamma}_l$ of the auto-covariance structure.  \\

An \textsc{The AutoRegressive (AR)} model, usually indexed with a parameter $p$ as $AR(p)$ models $x_t$ `a present observation' in terms of its $p$ immediate predecessors 
\begin{equation}
x_t = a + a_1x_{t-1} + a_2 x_{t-2} + \dots + a_p x_{t-p} + \epsilon_t \label{AR}
\end{equation}
where 
$a_0=1$, and $\sigma^2,a_1, a_2,..., a_p$ are the parameters of the model.  Our object is to estimate these parameters. Saturated models typically incorporate all of the X's while nested models incorporate a subset of the X's and a corresponding subset of the $a$'s; we refer to these respectively as $X^*$ and $a^*$. 
In vector terminology using the notation:
\[ X_{a:b}=(X_a,...,X_b); \quad X_0=1, \]
equations (\ref{AR}) can be written in the form: 
\begin{equation} X_{1:t} = A_t X_{0:t-1} + \epsilon \end{equation}
with $A_t=\begin{bmatrix} a_0=1 & 0 &... & 0 \\
a_0 & a_1 & ... & 0 \\ ..... & .... & ......& \end{bmatrix}$
 Yule-Walker equations dictate that, for the saturated model, parameters
$\mathbf{a}=(a_0=1,...,a_p)$ and covariances $\mathbf{\widehat{\gamma}}=(\widehat{\gamma}_0,...,\widehat{\gamma}_p)$ satisfy the equation:
\begin{eqnarray} 
\sigma^2 e_1 = \begin{bmatrix} \widehat{\gamma}_0 & \widehat{\gamma}_1 & .&.&. & \widehat{\gamma}_{p-1} \\
\widehat{\gamma}_1 & \widehat{\gamma}_0 &  .&.&.& \widehat{\gamma}_{p-2} \\
.&.&.&. & . & .  \\
\widehat{\gamma}_p & \widehat{\gamma}_{p-1} & .&.&. &
\widehat{\gamma}_0 
\end{bmatrix}  \mathbf{\phi}  \label{SMA}
\end{eqnarray}
which correspond to $p+1$ equations in $\sigma^2,\phi_1,...,\phi_p$ (i.e., $p+1$ unknowns).
The corresponding nested process equations take the form: 
\begin{eqnarray} 
\sigma^2 e_1 & = & \begin{bmatrix} 
 \widehat{\gamma}_0^* & \widehat{\gamma}_1^* & .&.&. & \widehat{\gamma}_{m-1}^* \\
\widehat{\gamma}_1^* & \widehat{\gamma}_0^* &  .&.&.& \widehat{\gamma}_{m-2}^* \\
.&.&.&. & . & .  \\
\widehat{\gamma}_m & \widehat{\gamma}_{m-1} & .&.&. &
\widehat{\gamma}_0 
\end{bmatrix}   \mathbf{\phi}^* \label{NM}
\end{eqnarray}
which again correspond to $p+1$ equations in $p+1$ unknowns.
The Yule-Walker equations (\ref{SMA}) and (\ref{NM}) show that the eigenvalues of the covariance matrix for the saturated model are smaller than those of the nested model.

The two-parameter \textsc{ARMA} model, $ARMA(p,q)$, combines $AR(p)$ and $MA(q)$ and takes the form:
\begin{equation}
 a_0 + a_1x_{t-1} + a_2 x_{t-2} + \dots + a_p x_{t-p} = \epsilon_t + b_1 \epsilon_{t-1} + \dots + b_q \epsilon_{t-q}. t=0,1,...
\end{equation} 
For the ${\rm ARMA}(p,q)$ model,
The Yule-Walker equations yield the equation:
\begin{eqnarray} 
\sigma_{MA,p}^2 e_1 = \begin{bmatrix} \widehat{\gamma}_0 & \widehat{\gamma}_1 & .&.&. & \widehat{\gamma}_{p-1} \\
\widehat{\gamma}_1 & \widehat{\gamma}_0 &  .&.&.& \widehat{\gamma}_{p-2} \\
.&.&.&. & . & .  \\
\widehat{\gamma}_p & \widehat{\gamma}_{p-1} & .&.&. &
\widehat{\gamma}_0 
\end{bmatrix}  \mathbf{\phi}  \label{ARMA}
\end{eqnarray}
which correspond to $p+1$ equations in $\sigma_{MA,p}^2,\phi_1,...,\phi_p$ (i.e, $p+1$ unknowns).
The moving average variance, $\sigma_{MA,p}$, takes the form
\[ \sigma_{MA,p}^2 = \sigma^2 (1 + b_1^2+...+b_q^2)  \]
which can be solved sequentially under the assumption that each of the variances are simulated (as was the case for the covariances).  The Yule-Walker equations show that the largest eigenvalue of the covariance matrix for the saturated model is larger than that of the nested model.

\subsection{An afterthought}\label{ssecacks} 
The first author noticed this obvious relation between bias and variance in generalized linear models and finds it rather surprising that either nobody has noticed it or, more likely, nobody has taken the trouble to spell it out.

Furthermore, the authors believe that a similar measure of `variance' is tied to complexity where other modeling frameworks are considered. This claim includes  neural networks, decision trees, various ensemble methods, and just about any modeling procedures where complexity is measurable.

\section*{Acknowledgments}
  The first author expresses his gratitude to his employer, Acronis SCS, and its parent company, Acronis for their steadfast support for research. Peter Shaw was supported in part by the Jiangsu province, China, 100 Talent project fund (BX2020100).

\bibliographystyle{unsrt}  
\bibliography{references}

\end{document}